\documentclass[conference]{IEEEtran}
\usepackage{times}

\usepackage[numbers,sort]{natbib}
\usepackage{multicol}
\usepackage{multirow}
\usepackage{hyperref}

\usepackage[nolist,nohyperlinks]{acronym}
\usepackage{graphicx}
\usepackage{subfig}
\usepackage{epstopdf}
\usepackage{epsfig}
\usepackage{algorithm}
\usepackage{algpseudocode}
\usepackage[disable]{todonotes} 
\usepackage{multirow}
\usepackage{flushend}
\usepackage{amsmath}
\usepackage[font=small,labelfont=bf]{caption}
\usepackage[binary-units=true]{SIunits}
\usepackage{verbatimbox}
\newcommand{\segmap}{\textit{SegMap}}

\begin{document}

\begin{minipage}{0.7\paperwidth}
\begin{center}
This paper has been accepted for publication in Robotics: Science and Systems.

\vspace{5mm}
DOI: 10.15607/RSS.2018.XIV.003


\vspace{10mm}
Please cite our work as:

\vspace{5mm}
R. Dub\'e, A. Cramariuc, D. Dugas, J. Nieto, R. Siegwart, and C. Cadena. ``SegMap: 3D Segment Mapping using Data-Driven Descriptors.'' Robotics: Science and Systems (RSS), 2018.
\end{center}

\vspace{10mm}
bibtex:
\begin{verbnobox}[\small]
@inproceedings{segmap2018,
  title      =  {{SegMap}: 3D Segment Mapping using Data-Driven Descriptors},
  author     =  {Dub{\'e}, Renaud and Cramariuc, Andrei and Dugas, Daniel and
                Nieto, Juan and Siegwart, Roland and Cadena, Cesar},
  booktitle  =  {Robotics: Science and Systems (RSS)},
  year       =  {2018}
}
\end{verbnobox}
\end{minipage}

\clearpage

\title{\segmap{}: 3D Segment Mapping using Data-Driven Descriptors}

\author{\authorblockN{{Renaud Dub\'e$^{*1,2}$, Andrei Cramariuc$^{*2}$, Daniel Dugas$^{2}$, Juan Nieto$^{2}$, Roland Siegwart$^{2}$, and Cesar Cadena$^{2}$}
\authorblockA{$^{1}$Sevensense Robotics AG ~ $^{2}$Autonomous Systems Lab, ETH, Zurich\\
Emails: renaud.dube@sevensense.ch and \{crandrei, dugasd, jnieto, rsiegwart, cesarc\}@ethz.ch \\
$^{*}$The authors contributed equally to this work.\\}
}
}


%

\maketitle

\begin{abstract}
%
When performing localization and mapping, working at the level of structure can be advantageous in terms of robustness to environmental changes and differences in illumination. 
%
This paper presents \segmap{}: a \textit{map representation} solution to the localization and mapping problem based on the extraction of segments in 3D point clouds.
%
%
In addition to facilitating the computationally intensive task of processing 3D point clouds, working at the level of segments addresses the data compression requirements of real-time single- and multi-robot systems.
While current methods extract descriptors for the single task of localization, \segmap{} leverages a data-driven descriptor in order to extract meaningful features that can also be used for reconstructing a dense 3D map of the environment and for extracting semantic information. 
%
This is particularly interesting for navigation tasks and for providing visual feedback to end-users such as robot operators, for example in search and rescue scenarios.
These capabilities are demonstrated in multiple urban driving and search and rescue experiments.
Our method leads to an increase of area under the ROC curve of 28.3\% over current state of the art using eigenvalue based features.
We also obtain very similar reconstruction capabilities to a model specifically trained for this task.
%
%
The \segmap{} implementation is available open-source along with easy to run demonstrations at \url{www.github.com/ethz-asl/segmap}.
\end{abstract}

\IEEEpeerreviewmaketitle

\section{Introduction}
Being a critical competency for mobile robotics, localization and mapping has been a well-studied topic over the last couple of decades~\cite{Cadena16tro-SLAMfuture}. 
In recent years, the importance of \ac{SLAM} has proven especially relevant in the context of applications with social impact, such as autonomous driving and disaster response. 
Although cameras and LiDARs are often used in conjunction due to their complementary nature~\cite{newman2009navigating}, the \ac{SLAM} problem for 3D LiDAR point clouds still poses open challenges.
Moreover, LiDAR based systems rely on structure which can be more consistent than visual appearance across seasons and daylight changes.
Despite recent developments, a number of important capabilities are still lacking in many existing 3D LiDAR SLAM frameworks. 
Perhaps most notably, this includes the absence of global data associations (place recognitions) from almost all 3D LiDAR based \ac{SLAM} implementations, while contrastingly being a well-studied problem in visual \ac{SLAM}~\cite{lowryvisual}. 

%
%
%
\begin{figure}
\centering
\includegraphics[width=1.0\columnwidth]{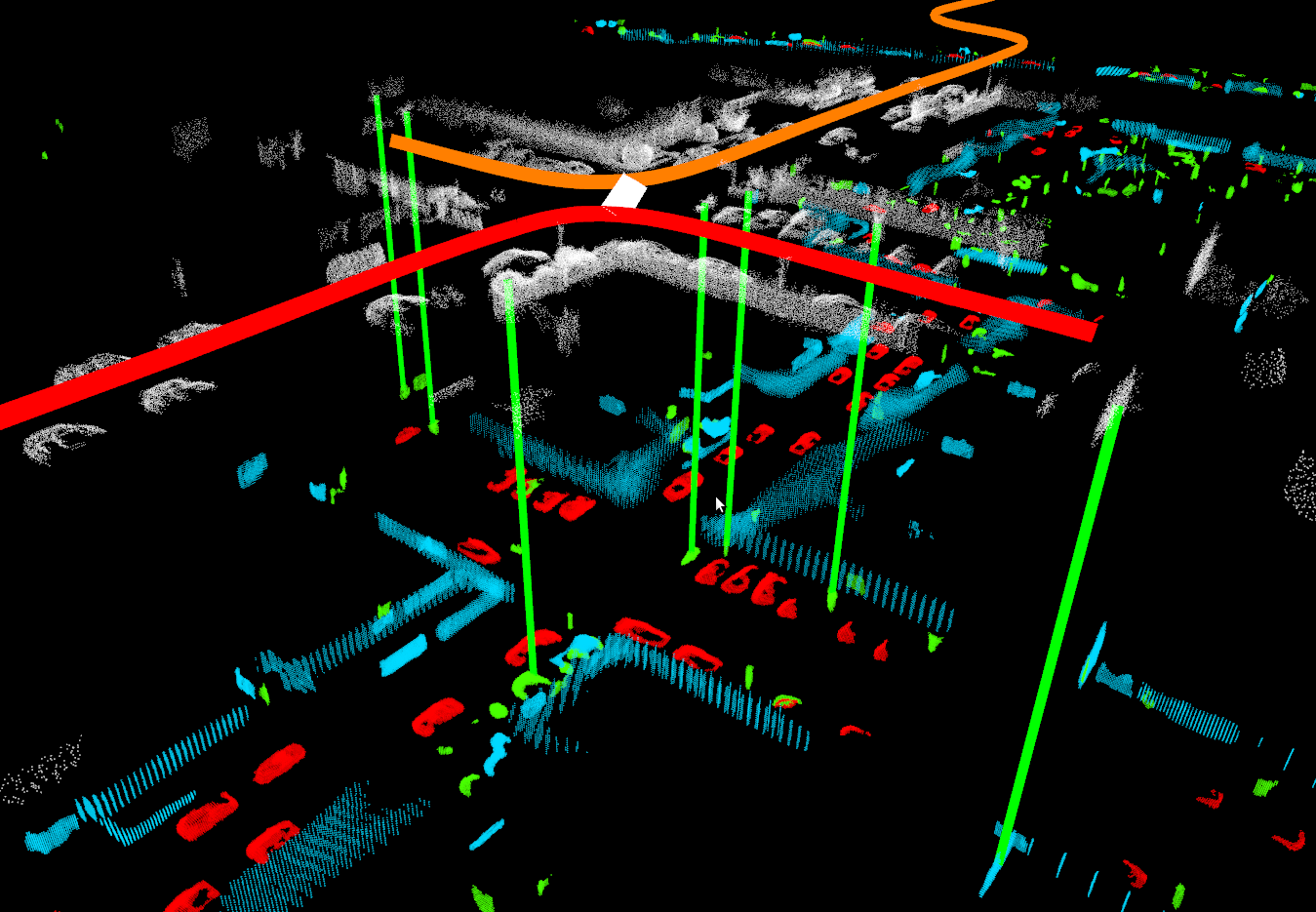}
\vspace{-5mm}
\caption{An illustration of the \segmap{} approach\protect\footnotemark.
The red and orange lines represent two robots driving simultaneously in opposite directions through an intersection.
In white we show the local segments extracted from the robots' vicinity and characterized using our compact data-driven descriptor.
Correspondences are then made with the target segments, resulting in successful localizations depicted with green vertical lines.
A reconstruction of the target segments is illustrated below, where colors represent semantic information (cars in red, buildings in light blue, and others in green), all possible by leveraging the same compact representation.
We take advantage of the semantic information by performing localization only against static objects, adding robustness against dynamic changes.}
\label{fig_teaser}
\vspace{-7mm}
\end{figure}

%
%
%
%
%
%
%

This paper presents \segmap{}: a unified approach for \textit{map representation} in the localization and mapping problem for 3D LiDAR point clouds.
The \segmap{} approach is formed on the basis of partitioning point clouds into sets of descriptive segments~\cite{dube2017segmatch}, as illustrated in Figure~\ref{fig_teaser}.
Segments are obtained using clustering techniques which are able to repeatedly form similar partitions of the point cloud.
The resulting segments provide the means for compact, yet discriminative features to represent the environment efficiently.
Global data associations are identified by segment descriptor retrieval, made possible by the repeatable and descriptive nature of segment based features. 
%
%
%
%
The use of segment based features facilitates low computational, memory and bandwidth requirements, and therefore makes the approach appropriate for real-time use in both multi-robot and long-term applications.
Moreover, as segments typically represent meaningful and distinct elements that make up the environment, a scene can be effectively summarized by a handful of compact feature descriptors.\footnotetext{A video demonstration is available at \url{https://youtu.be/CMk4w4eRobg}.}

Previous work on segment based localization considered hand-crafted features and provided a sparse representation\cite{dube2017segmatch}.
These features lack the ability to generalize to different environments and only offer limited insights into the underlying 3D structure.
In this work, we overcome these shortcomings by introducing a novel data-driven segment descriptor which can offer high retrieval performances, even under variations in point of view.
As depicted in Figure~\ref{fig_teaser}, these descriptors can be decoded in order to generate 3D reconstructions.
These can be used by robots for navigating around obstacles and displayed to remote operators for improved situation awareness.
Moreover, we show that semantic information can be extracted by performing classification in the descriptor space.
This information can for example lead to increased robustness to changes in the environment.

To the best of our knowledge, this is the first work on robot localization proposing to reuse the extracted features for reconstructing environments in three dimensions and for extracting semantic information.
This reconstruction is, in our opinion, a very interesting capability for real-world, large-scale applications with limited memory and communication bandwidth.
To summarize, this paper presents the following contributions:
\begin{itemize}
\item A novel data-driven 3D segment descriptor achieving increased localization performance.
\item A technique for reconstructing the environment based on the same compact features used for localization.
\item An extensive evaluation of the \segmap{} approach using real-world, multi-robot automotive and disaster scenario datasets.
\end{itemize}

The remainder of the paper is structured as follows: Section~\ref{sec:related_work} provides an overview of the related work in the fields of localization and machine learning based descriptors for 3D point clouds.
The \segmap{} approach and our novel descriptor enabling environment reconstruction are detailed in Section~\ref{sec:segmap} and Section~\ref{sec:descriptor}.
The method is evaluated in Section~\ref{sec:experiments}, and Section~\ref{sec:conclusion} finally concludes with a short discussion.

\section{RELATED WORK}
\label{sec:related_work}

An overview of the related work on localization in 3D point clouds was presented in~\cite{Cadena16tro-SLAMfuture} and~\cite{dube2017segmatch}.
In this section, we review learning based techniques with applications to 3D points clouds. 

In recent years, \acp{CNN} have become the state of the art method for generating learning based descriptors, due to their ability to find complex patterns in data~\cite{krizhevsky2012imagenet}. 
For 3D point clouds methods based on \acp{CNN} achieve impressive performances in applications such as object detection~\cite{engelcke2017vote3deep, maturana2015voxnet, riegler2016octnet, livehicle, wu20153d, wohlhart2015learning, qi2016pointnet}, semantic segmentation~\cite{riegler2016octnet, livehicle, qi2016pointnet, tchapmi2017segcloud}, 
and 3D object generation~\cite{wu2016learning}.

Recently, a handful of works proposing the use of \acp{CNN} for localization in 3D point clouds have started to appear~\cite{zeng20163dmatch, elbaz20173d}. 
First, \citet{zeng20163dmatch} propose extracting data-driven 3D keypoint descriptors (3DMatch) which are robust to changes in point of view. 
Although impressive retrieval performances are demonstrated using an RGB--D sensor in indoor environments, it is not clear whether this method is applicable in real-time in large-scale outdoor environments.
%
%
\citet{elbaz20173d} propose describing local subsets of points using a deep neural network autoencoder. 
The authors state that the implementation has not been optimized for real-time operation and no timings have been provided.
Contrastingly, our work presents a data-driven segment based localization method that can operate in real-time and that allows map reconstruction and semantic extraction capabilities. 


To achieve this reconstruction capability, the architecture of our descriptor was inspired by autoencoders in which an encoder network compresses the input to a small dimensional representation, and a  decoder network attempts to decompress the representation back into the original input.
%
%
The compressed representation can be used as a descriptor for performing 3D object classification~\cite{brock2016generative}.
\citet{brock2016generative}~also present successful results using variational autoencoders for reconstructing voxelized 3D data.
%
Different configurations of encoding and decoding networks have also been proposed for achieving localization and for reconstructing and completing 3D shapes and environments~\cite{guizilini2017learning, dai2016shape, varley2016shape, ricao2017compressed,elbaz20173d,schonberger2017semantic}.

While autoencoders present the interesting opportunity of simultaneously accomplishing both compression and feature extraction tasks, optimal performance at both is not guaranteed.
As will be shown in Section~\ref{ssec:retrieval_performance}, these two tasks can have conflicting goals when robustness to changes in point of view is desired.
In this work, we combine the advantages of the encoding-decoding architecture of autoencoders with a technique proposed by~\citet{parkhi2015deep}.
The authors address the face recognition problem by first training a \ac{CNN} to classify people in a training set and afterwards use the second to last layer as a descriptor for new faces. 
This classification based method proved to be the best in our previous work where we evaluated three training techniques for achieving better segment descriptor retrieval performances~\cite{cramariuc2017}.
Other alternative training techniques include for example the use of contrastive loss~\cite{bromley1994signature} or triplet loss~\cite{weinberger2006distance}.
We use the resulting segment descriptors in the context of \ac{SLAM} to achieve better performance, as well as significantly compressed maps that can easily be stored, shared, and reconstructed.

%
%
%
 
%
%


\section{The \segmap{} approach}
\label{sec:segmap}
This section presents our \segmap{} approach to localization and mapping in 3D point clouds.
It is composed of five core modules: segment extraction, description, localization, map reconstruction, and semantics extraction.
These modules are detailed in this section and together allow single and multi-robot systems to create a powerful unified representation which can conveniently be communicated. 

%
%

\textbf{Segmentation}\hspace{2pt} The stream of point clouds generated by a 3D sensor is first accumulated in a dynamic voxel grid\footnote{In our experiments, LiDAR-odometry is estimated by performing scan registration using \ac{ICP}~\cite{dube2017multirobot}. In future work, it would be interesting to combine the \segmap{} approach with other LiDAR based odometry techniques~\cite{zhang2014loam, Bosse2009a}}.
Point cloud segments are then extracted in a section of radius $R$ around the robot.
An incremental region growing algorithm is used to efficiently grow segments by using only newly active voxels as seeds~\cite{dube2018incremental}.
This results in a handful of local segments, which are each associated to a set of past observations i.e. ${S_i=\{s_1, s_2, \ldots, s_n \}}$. 
Each observation $s_j \in S_i$ is a 3D point cloud representing a snapshot of the segment as points are added to it.
Note that $s_n$ represents the latest observation of a segment and is considered \textit{complete} when no further measurements are made, e.g. when the robot has moved away.

\textbf{Description}\hspace{2pt} Compact features are then extracted from these 3D segment point clouds using the data-driven descriptor presented in Section~\ref{sec:descriptor}.
A global segment map is created online by accumulating the segment centroids and descriptors.
In order for the global map to most accurately represent the latest state of the world, we only keep the descriptor associated with the last and most complete observation. 

\textbf{Localization}\hspace{2pt} In the next step, candidate correspondences are identified between global and local segments using \ac{k-NN} in feature space. 
Localization is finally performed by verifying the candidate correspondences for geometric consistency on the basis of the segment centroids.
In the experiments presented in Section~\ref{final_experiment}, this is achieved using an incremental recognition strategy based on partitioning and caching of geometric consistencies~\cite{dube2018incremental}.
When a geometrically consistent set of correspondence is identified, a 6 \ac{DoF} transformation between the local and global maps is estimated.
This transformation is fed to an incremental pose-graph \ac{SLAM} solver which in turn estimates, in real-time, the trajectories of all robots~\cite{dube2017multirobot}.

\textbf{Reconstruction \& Semantics}\hspace{2pt} The compressed representation can at any time be used to reconstruct a map and to extract semantic information.
Thanks to the compactness of the \segmap{} descriptor which can conveniently be transmitted over wireless networks with limited bandwidth, any agent in the network can reconstruct and leverage this 3D information.
On the other hand, the semantic information can for example be used to discern between static and dynamic objects which can improve the robustness of the localization.


\section{The \segmap{} Descriptor}
\label{sec:descriptor}
In this section we present our main contribution: a data-driven descriptor for 3D segment point clouds which allows for localization, map reconstruction and semantic extraction.
The descriptor extractor's architecture and the processing steps for inputting the point clouds to the network are introduced.
We then describe our technique for training this descriptor to accomplish both tasks of segment retrieval and map reconstruction.
We finally show how the descriptor can further be used to extract semantic information from the point cloud.

\subsection{Descriptor extractor architecture}
\begin{figure*}
\centering
\includegraphics[width=7in]{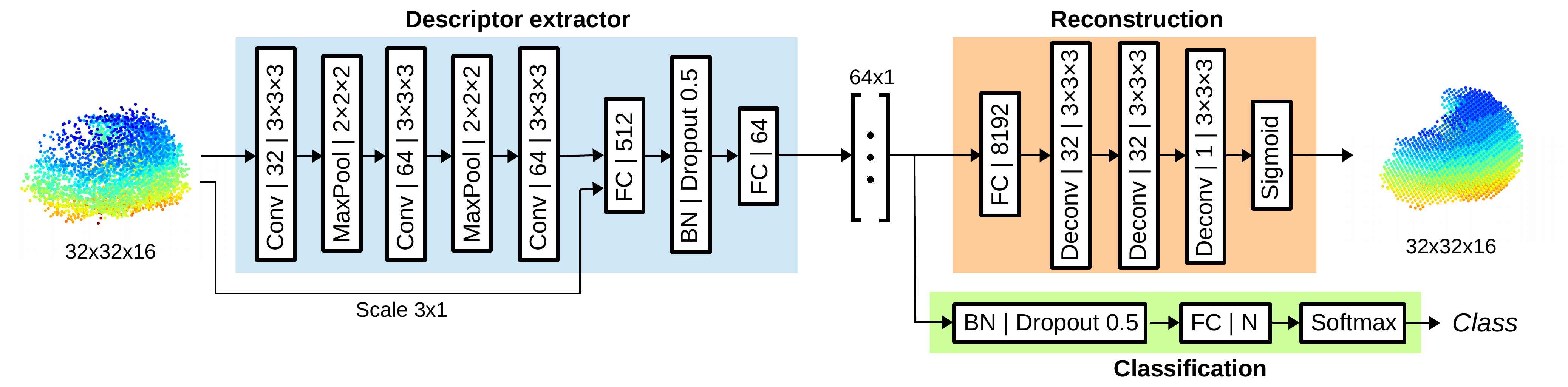}
\vspace{-1mm}
\caption{
The descriptor extractor is composed of three convolutional and two fully connected layers.
The 3D segments are compressed in a representation of dimension  $64\times1$ which can be used for localization, map reconstruction and semantic extraction.
Right of the descriptor we illustrate the classification and reconstruction layers which are used for training.
In the diagram the convolutional (Conv), deconvolutional (Deconv), fully connected (FC) and batch normalization (BN) layers are abbreviated respectively.
As parameters the Conv and Deconv layers have the number of filters and their sizes, FC layers have the number of nodes, max pool layers have the size of the pooling operation, and dropout layers have the ratio of values to drop.
Unless otherwise specified, \ac{ReLU} activation functions are used for all layers.
}
\label{fig:descriptor_architecture}
\vspace{-6mm}
\end{figure*}

The architecture of the descriptor extractor is presented in Fig.~\ref{fig:descriptor_architecture}.
Its input is a 3D binary voxel grid of fixed dimension $32\times32\times16$ which was determined empirically to offer a good balance between descriptiveness and the size of the network.
The description part of the \ac{CNN} is composed of three 3D convolutional layers with max pool layers placed in between and two fully connected layers.
Unless otherwise specified,\ac{ReLU} activation functions are used for all layers.
The original scale of the input segment is passed as an additional parameter to the first fully connected layer to increase robustness to voxelization at different aspect ratios.
The descriptor is obtained by taking the activations of the extractor's last fully connected layer.
This architecture was found by grid searching through various parameters.

\subsection{Segment alignment and scaling}
\label{sec:alignment}
A pre-processing stage is required in order to input the 3D segment point clouds for description.
First, an alignment step is applied such that segments extracted from the same objects are similarly presented to the descriptor network.
This is performed with the assumption that the $z$-axis is roughly aligned with gravity and by applying a 2D Principal Components Analysis (PCA) of all points located within a segment.
The segment is then rotated so that the $x$-axis of its frame of reference aligns with the eigenvector corresponding to the largest eigenvalue.
We choose to solve the ambiguity in direction by rotating the segment so that the lower half section along the $y$-axis of  its frame of reference contains the highest number of points.
%
From the multiple alignment strategies we evaluated, the presented strategy worked best.

The network's input voxel grid is applied to the segment so that its center corresponds to the centroid of the aligned segment.
By default the voxels have minimum side lengths of \unit{0.1}{\meter}.
These can individually be increased to exactly fit segments having one or more larger dimension than the grid.
Whereas maintaining the aspect ratio while scaling can potentially offer better retrieval performance, this individual scaling with a minimum side length better avoids large errors caused by aliasing. 
We also found that this scaling method offers the best reconstruction performance, with only a minimal impact on the retrieval performance when the original scale of the segments is passed as a parameter to the network.

\subsection{Training the \segmap{} descriptor}
\label{ssec:training}
In order to achieve both a high retrieval performance and reconstruction capabilities, we propose a customized learning technique.
The two desired objectives are imposed on the network by the \textit{softmax cross entropy loss} $L_c$ for retrieval and the reconstruction loss $L_r$.
We propose to simultaneously apply both losses to the descriptor and to this end define a combined loss function $L$ which merges the contributions of both objectives:
\begin{equation}
L = L_c + \alpha L_r
\end{equation}
where the parameter $\alpha$ weighs the relative importance of the two losses.
The value $\alpha=200$ was empirically found to not significantly impact the performance of the combined network, as opposed to training separately with either of the losses. 
Weights are initialized based on Xavier's initialization method \cite{glorot2010understanding} and trained using the \ac{ADAM} optimizer~\cite{adam} with a learning rate of $10^{-4}$.
In comparison to \ac{SGD}, \ac{ADAM} maintains separate learning rates for each network parameter, which facilitates training the network with two separate objectives simultaneously.
Regularization is achieved using dropout~\cite{srivastava2014dropout} and batch normalization~\cite{ioffe2015batch}.


\textbf{Classification loss} $\boldsymbol{L_c}$ \hspace{3pt}For training the descriptor to achieve better retrieval performance, we use a learning technique similar to the \textit{N-ways classification problem} proposed by~\citet{parkhi2015deep}. 
Specifically, we organize the training data into $N$ classes where each class contains all observations of a segment or of multiple segments that belong to the same object or environment part.
%
%
Note that these classes are solely used for training the descriptor and are not related to the semantics presented in Section~\ref{ssec:training_semantics}.
As seen in Fig~\ref{fig:descriptor_architecture}, we then append a classification layer to the descriptor and teach the network to associate a score to each of the $N$ predictors for each segment sample.
These scores are compared to the true class labels using \textit{softmax cross entropy loss}:
\begin{equation}
L_c = -\sum^{N}_{i=1} y_i\log{\frac{e^{l_i}}{\sum^{N}_{k=1} e^{l_k}}}
\end{equation}
where $y$ is the one hot encoded vector of the true class labels and $l$ is the layer output.

\begin{figure}
\centering
\includegraphics[width = 0.75\columnwidth]{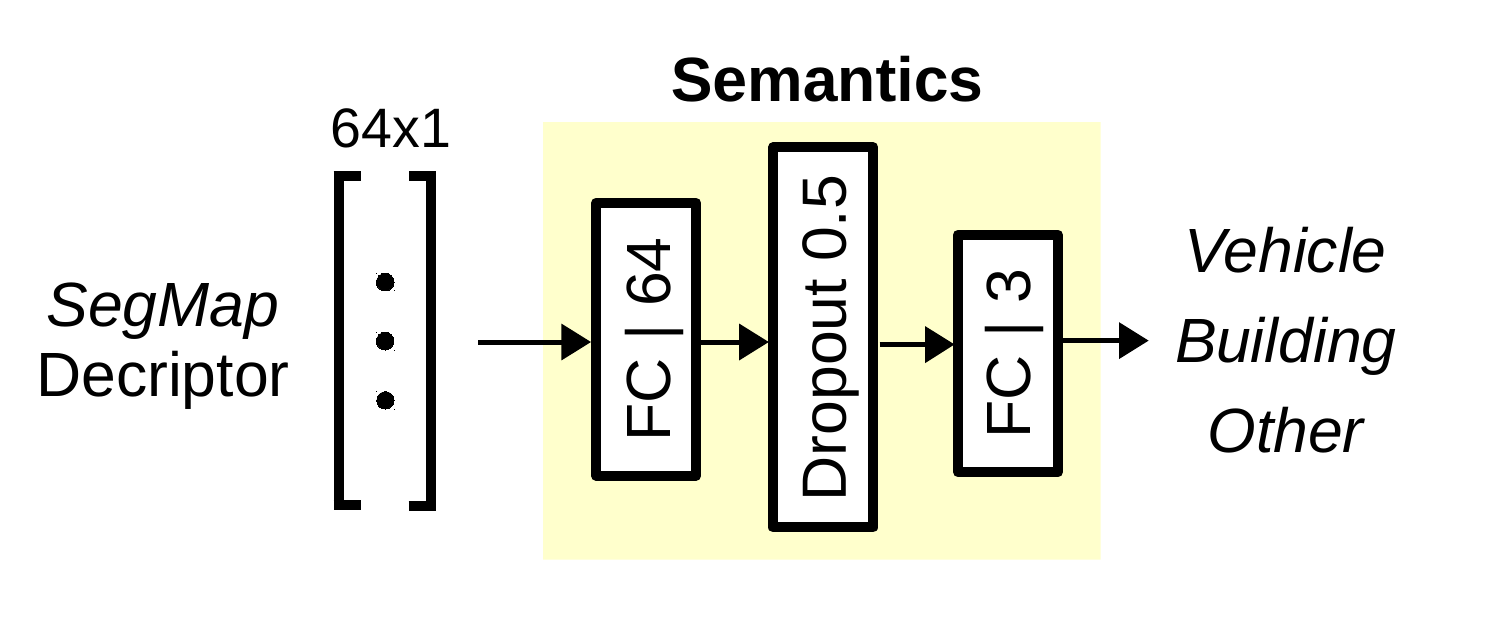}
\vspace{-3mm}
\caption{A simple fully connected network that can be appended to the \segmap{} descriptor (depicted in Fig.~\ref{fig:descriptor_architecture}) in order to extract semantic information.
In our experiments, we train this network to distinguish between vehicles, buildings, and other objects.
}
\label{fig_semantics}
\vspace{-6mm}
\end{figure}

Given a large number of classes and a small descriptor dimensionality, the network is forced to learn descriptors that better generalize and prevent overfitting to specific segment samples. 
Note that when deploying the system in a new environment the classification layer is removed, as the classes are no longer relevant.
The activations of the previous fully connected layer are then used as a descriptor for segment retrieval through \ac{k-NN}.

\textbf{Reconstruction loss} $\boldsymbol{L_r}$ \hspace{3pt}As depicted in Fig.~\ref{fig:descriptor_architecture}, map reconstruction is achieved by appending a decoder network and training it simultaneously with the descriptor extractor and classification layer.
This decoder is composed of one fully connected and three deconvolutional layers with a final sigmoid output.
Note that no weights are shared between the descriptor and the decoder networks.
Furthermore, only the descriptor extraction needs to be run in real-time on the robotic platforms, whereas the decoding part can be executed any time a reconstruction is desired.

As proposed by~\citet{brock2016generative}, we use a specialized form of the \textit{binary cross entropy loss}, which we denote by $L_r$:
\begin{equation}
L_r = -\sum_{x,y,z}\gamma t_{xyz} \log(o_{xyz}) + (1 - \gamma) (1 - t_{xyz}) \log(1 - o_{xyz})
\end{equation}
where $t$ and $o$ respectively represent the target segment and the network's output and $\gamma$ is a hyperparameter which weighs the relative importance of false positives and false negatives.
This parameter addresses the fact that only a minority of voxels are activated in the voxel grid.
In our experiments, the voxel grids used for training were on average only 3\% occupied and we found $\gamma = 0.9$ to yield good results. 

\begin{figure*}
  \centering
  \includegraphics[width=7.1in]{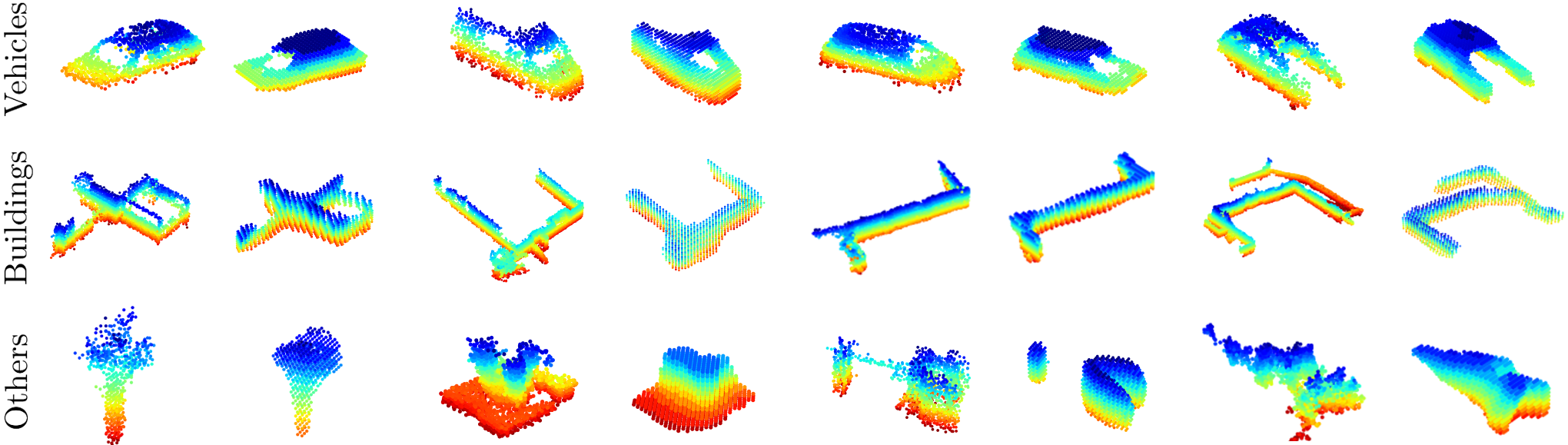}
  \vspace{-2mm}
  \caption{An illustration of the \segmap{} reconstruction capabilities.
  The segments are extracted from sequence 00 of the KITTI dataset and represent, from top to bottom respectively, vehicles, buildings, and other objects. 
  For each segment pair, the reconstruction is shown at the right of the original.
  The network manages to accurately reconstruct the segments despite the high compression to only $64$ values.
  Note that the voxelization effect is more visible on buildings as larger segments necessitate larger voxels to keep the input dimension fixed.}
  \label{fig:reconstructions}
  \vspace{-6mm}
\end{figure*}

\subsection{Knowledge transfer for semantic extraction}
\label{ssec:training_semantics}
As can be observed from Fig.~\ref{fig_teaser}, segments extracted by the \segmap{} approach for localization and map reconstruction often represent objects or parts of objects.
It is therefore possible to assign semantic labels to these segments and use this information to improve the performance of the localization process.
As depicted in Fig.~\ref{fig_semantics}, we transfer the knowledge embedded in our compact descriptor by training a semantic extraction network on top of it.
This last network is trained with labelled data using the \textit{softmax cross entropy loss} and by freezing the weights of the descriptor network.

In this work, we choose to train this network to distinguish between three different semantic classes: \textit{vehicles}, \textit{buildings}, and \textit{others}.
Section~\ref{final_experiment} shows that this information can be used to increase the robustness of the localization algorithm to changes in the environment and to yield smaller map sizes.
This is achieved by rejecting segments associated with potentially dynamic objects, such as vehicles, from the list of segment candidates.

\section{EXPERIMENTS}
\label{sec:experiments}
This section presents the experimental validation of our approach.
We first present a procedure for generating training data and detail the performances of the \segmap{} descriptor for localization, reconstruction and semantics extraction.
We finally demonstrate the performance of the \segmap{} approach in two large scale experiments.

\subsection{Experiment setup and implementation}
All experiments were performed on a system equipped with an Intel i7-6700K processor, and a Nvidia GeForce GTX 980 Ti GPU.
The \ac{CNN} models were developed and executed in real-time using the TensorFlow library.
The library $libnabo$ is used for descriptor retrieval with fast \ac{k-NN} search in low dimensional space~\cite{elsebergcomparison}.
The incremental optimization back-end is based on the iSAM2 implementation~\cite{Kaess2012a}.

\subsection{Baselines}
\label{ssec:baselines}
In the following experiments, our \segmap{} descriptor is compared with eigenvalue based point cloud features~\cite{weinmann2014semantic} and with a \ac{CNN} trained specifically for compressing and reconstructing segment point clouds.
This purely autoencoder model has the exact same architecture presented in Fig.~\ref{fig:descriptor_architecture}.
The single difference is that it is trained solely for reconstructing segment point clouds, i.e. by using only the reconstruction loss $L_r$.
We will refer to these two baselines as Eigen for the eigenvalue based features and AE for the autoencoder model.
For reference, previous work proposed to describe 3D segments using the ensemble of shape histograms  \cite{wohlkinger2011ensemble, dube2017segmatch}.
However, this descriptor was not included in our evaluation as its high dimensionality is not well suited to our goals of map compression and efficient \ac{k-NN} retrieval in large maps.

\subsection{Training data}
\label{ssec:experiments_learning}
The \segmap{} descriptor is trained using real-world data from the KITTI odometry dataset~\cite{geiger2012we}.
Sequences 05 and 06 are used for generating training data whereas sequence 00 is solely used for evaluating the descriptor performances.
For each sequence, segments are extracted using an incremental Euclidean distance based region growing technique~\cite{dube2018incremental}.
This algorithm extracts point clouds representing parts of objects or buildings which are separated after removing the ground plane (see Fig.~\ref{fig:reconstructions}).
The training data is filtered by removing segments with too few observations, or training classes (as described in Section~\ref{ssec:training}) with too few samples. 
In this manner, 3300, 1750, and 810 segments are respectively generated from sequences 00, 05, and 06 with an average of 12 observations per segment over the whole dataset.

\subsubsection{Data augmentation}
To further increase robustness by reducing sensitivity to rotation and point of view changes in the descriptor extraction process, the dataset is augmented by using multiple copies of the same segment rotated at different angles to the alignment described in \ref{sec:alignment}.
Also, in order to simulate the effect of occlusion we generate artificial copies of each segment by removing all points which fall on one side of a randomly generated slicing plane that does not remove more than $50$\% of the points.
Note that these two data augmentation steps are performed prior to voxelization. 
%

\subsubsection{Ground-truth generation}
\label{ssec:ground_truth}
In the following step, we use GPS information in order to identify ground truth correspondences between segments extracted in areas where the vehicle performed multiple visits.
Only segment pairs with a maximum distance between their centroids of \unit{3.0}{\meter} are considered.
We compute the 3D convex hull of each segment observation $s_1$ and $s_2$ and create a correspondence when the following condition, inspired from the Jaccard index, holds:
\begin{equation}
\frac{\text{Volume}(\text{Conv}(s_1) \cap \text{Conv}(s_2))}{\text{Volume}(\text{Conv}(s_1) \cup \text{Conv}(s_2))} {\geq} p
\end{equation}
In our experiments we found $p = 0.3$ to generate a sufficient number of correspondences while preventing false labelling.
The procedure is performed on sequences 00, 05, and 06, generating 150, 260, and 320 ground truth correspondences respectively.
We use two-thirds of the correspondences for augmenting the training data and one-third for creating validation samples.
Finally, the ground-truth correspondences extracted from sequence 00 are used in Section~\ref{ssec:retrieval_performance} for evaluating the retrieval performances.

\subsection{Training the models}
\label{training_models}
\begin{figure}
  \centering
  \includegraphics[width=1.0\columnwidth]{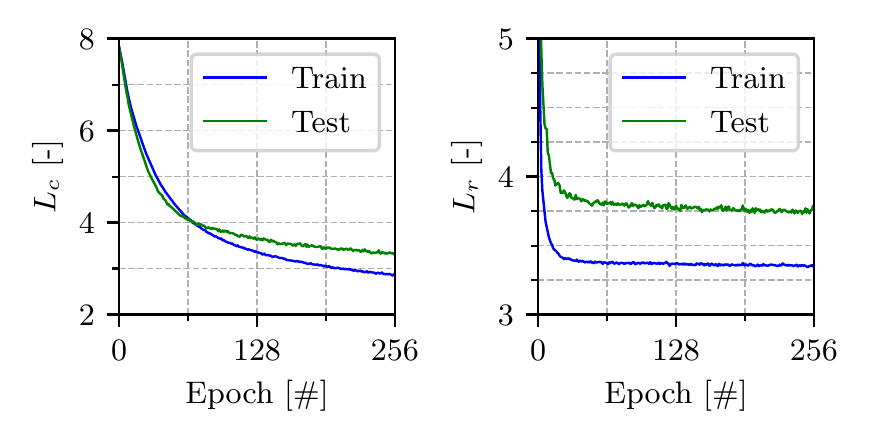}
  \vspace{-7mm}
  \caption{The classification loss $L_c$ (left) and the reconstruction loss $L_r$ (right) when training the descriptor extractor along with the reconstruction and classification networks.
The depicted reconstruction loss has already been scaled by $\alpha$.}
  \label{fig:loss}
  \vspace{-6mm}
\end{figure}

The descriptor extractor and the decoding part of the reconstruction network are trained using all segments extracted from drive 05 and 06.
Training lasts three to four hours on the GPU and produces the classification and scaled reconstruction losses depicted in Fig.~\ref{fig:loss}.
The total loss of the model is the sum of the two losses as describe in Section~\ref{ssec:training}.
We note that for classification the validation loss follows the training loss before converging towards a corresponding accuracy of 41\% and 43\% respectively.
In other words, 41\% of the validation samples were correctly assigned to one of the $N=2500$ classes.
This accuracy is expected given the large quantity of classes and the challenging task of discerning between multiple training samples with similar semantic meaning but few distinctive features, e.g. flat walls.
Note that we achieve very similar classification losses when training with and without the $L_r$ component of the loss.


%
%



\subsection{Retrieval performance}
\label{ssec:retrieval_performance}
\begin{figure}
  \vspace{-1mm}
  \centering
  \includegraphics{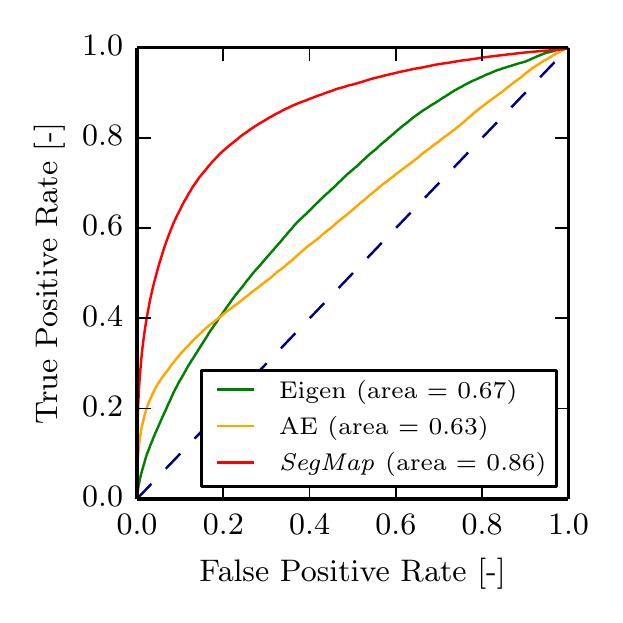}
  \vspace{-4mm}
  \caption{\ac{ROC} curves for the descriptors considered in this work.
  This evaluation is performed using ground-truth correspondences extracted from sequence 00 of the KITTI odometry dataset~\cite{geiger2012we}.}
  \label{fig:roc}
  \vspace{-6mm}
\end{figure}

The retrieval performances of the \segmap{}, eigenvalue based, and autoencoder descriptors are depicted in Fig~\ref{fig:roc}.
The \ac{ROC} curves are obtained by generating $45$M labelled pairs of segment descriptors from sequence 00 of the KITTI odometry dataset~\cite{geiger2012we}.
For each ground-truth correspondence, a positive sample is created for each possible segment observation pair. 
For each positive sample a thousand negative samples are generated by randomly sampling segment pairs whose centroids are further than \unit{20}{\meter} apart.
The positive to negative sample ratio is representative of our localization problem given that a map created from KITTI sequence 00 contains around a thousand segments.
The \ac{ROC} curves are finally obtained by varying the threshold applied on the $l^2$ distance between the two segment descriptors.  

\begin{figure}
  \vspace{-3mm}
  \centering
  \includegraphics{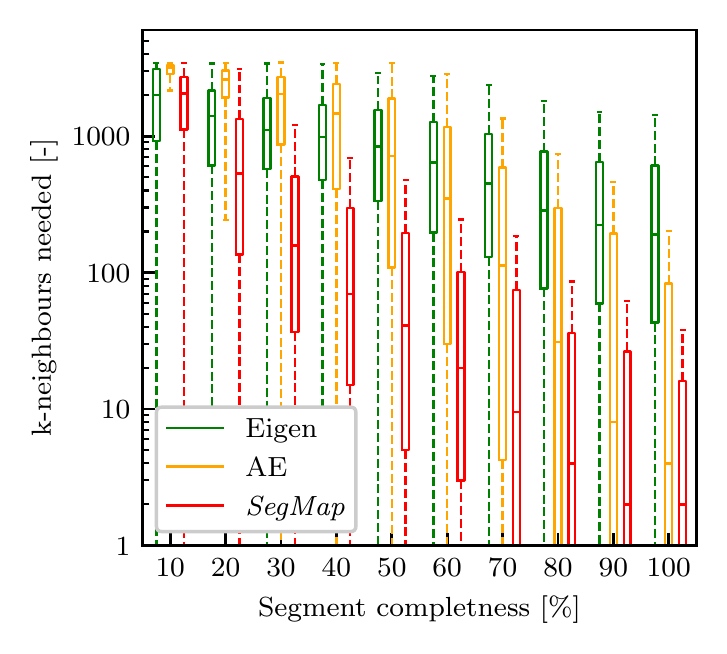}
  \vspace{-4mm}
  \caption{
This figure presents how \textit{quickly} descriptors extracted from incrementally grown segments contain relevant information that can be used for localization. 
The x-axis represents the growing status of a segment until all its measurements have been accumulated (here termed \textit{complete}, see Section~\ref{sec:segmap}).
The log-scaled y-axis represents how many neighbours in the target map need to be considered in order to include the correct target segment (the lower the better).
The \segmap{} descriptor offers one order of magnitude better retrieval performance for over 40\% of the growing process.
%
  }
  \label{fig:acc_relative_to_size}
  \vspace{-6mm}
\end{figure}

As introduced in Section~\ref{sec:segmap}, correspondences are made between segments from the local and global maps by using k-NN retrieval in feature space.
In order to avoid false localizations, the aim is to reduce the number $k$ of neighbours that need to be considered.
Therefore, as a segment grows with time, it is critical that its descriptor converges as quickly as possible towards the descriptor of the corresponding segment in the target map, which in our case is extracted from the last and most \textit{complete} observation (see Section~\ref{sec:segmap}). 
This behaviour is evaluated in Fig.~\ref{fig:acc_relative_to_size} which relates the number of neighbours which need to be considered to find the correct association as a function of segment completeness.  
We note that the \segmap{} descriptor offers the best retrieval performance at every stage of the growing process.
In practice this is important since it allows closing challenging loops such as the one presented in Fig.~\ref{fig_teaser}.
Interestingly, the autoencoder has the worst performance at the early growing stages whereas good performance is observed at later stages.
This is in accordance with the capacity of autoencoders to precisely describe the geometry of a segment, without explicitly  aiming at gaining robustness to changes in point of view.


\subsection{Reconstruction performance}
In addition to offering high retrieval performances, the \segmap{} descriptor allows us to reconstruct 3D maps using the decoding \ac{CNN} described in Section~\ref{ssec:training}.
Some examples of the resulting reconstructions are illustrated in Fig~\ref{fig:reconstructions}, for various objects captured during sequence 00 of the KITTI odometry dataset.
Experiments done at a larger scale are presented in  Fig.~\ref{fig:reconstruction_sar} where buildings of a powerplant and a foundry are reconstructed by fusing data from multiple sensors.
Overall, the reconstructions are well recognizable despite the high compression ratio.
We note that the quantization error resulting from the voxelization step mostly affects larger segments that have been downscaled to fit into the voxel grid.
To avoid this problem, one could consider different network structures which do not rely on voxelization, for example the one proposed by~\citet{qi2016pointnet}.

Since most segments only sparsely model real-world surfaces, they occupy on average only 3\% of the voxel grid.
To obtain a visually relevant comparison metric, we calculate for both the original segment and its reconstruction the ratio of points having a corresponding point in the other segment, within a distance of one voxel.
The tolerance of one voxel means that the shape of the original segment must be preserved while not focusing on reconstructing each individual point.
Results calculated for different descriptor sizes are presented in Table~\ref{tbl:stats_ae}, in comparison with the purely reconstruction focused baseline detailed in Sec.~\ref{ssec:baselines}.
The \segmap{} descriptor with a size of 64 has on average 91\% correspondences between the points in the original and reconstructed segments, and is only slightly outperformed by the AE baseline.
Contrastingly, the significantly higher retrieval performances of the \segmap{} descriptor makes it a clear all-rounder choice for achieving both localization and map reconstruction.

\subsection{Semantic extraction performance}
For training the semantic extractor network (Fig.~\ref{fig_semantics}), we manually labelled the last observation of all 1750 segments extracted from KITTI sequence 05.
The labels are then propagated to each observation of a segment for a total of 20k labelled segment observations.
We use 70\% of the samples for training the network and 30\% for validation.
Given the low complexity of the semantic extraction network and the small amount of labelled samples, training takes only a few minutes.
We achieve an accuracy of 89\% and 85\% on the training and validation data respectively.
Note that our goal is not to improve over other semantic extraction methods~\cite{livehicle, qi2016pointnet}, but rather to illustrate that our compressed representation can additionally be used for gaining robustness to dynamic changes and for reducing the map size (Section~\ref{ssec_kitti_slam}). 

\subsection{Large scale experiments}
\label{final_experiment}
We evaluate the \segmap{} approach on three large-scale multi-robot experiments: one in an urban-driving environment and two in search and rescue scenarios.
In both indoor and outdoor scenarios we use the same model which was trained on the KITTI sequences 05 and 06 as described in Section~\ref{training_models}.
In all experiments, we consider $40$ neighbours when performing segment retrieval and require a minimum of $7$ correspondences which are altogether geometrically consistent to output a localization.
These parameters were chosen empirically using KITTI sequence 00 and the information presented in Fig.~\ref{fig:roc}~and~\ref{fig:acc_relative_to_size} as a reference.

The experiments are run on one single machine, with a multi-thread approach to simulating a centralized system.
One thread per robot accumulates the 3D measurements, extracting segments, and performing the descriptor extraction.
The descriptors are transmitted to a separate thread which localizes the robots, through descriptor retrieval and geometric verification, and runs the pose-graph optimization. 
In all experiments, sufficient global associations need to be made, in real-time, for linking the trajectories and merging the maps. 
Moreover in a centralized setup it can be crucial to limit the transmitted data over a wireless network with potentially limited bandwidth. 


\begin{table}
\fontsize{8}{10}\selectfont
  \renewcommand{\arraystretch}{1.2}
  \centering
  \caption{Average ratio of corresponding points within one voxel distance between original and reconstructed segments. 
  Statistics for \segmap{} and the AE baseline using different descriptor sizes.}
  \label{tbl:stats_ae}
  \begin{tabular}{c|c|c|c|c|} \cline{2-5}
    \quad & \multicolumn{4}{|c|}{\textbf{Descriptor size}} \\ 
    \quad & \multicolumn{1}{|c}{16} & \multicolumn{1}{c}{32} & \multicolumn{1}{c}{64} & \multicolumn{1}{c|}{128} \\ \hline
    \multicolumn{1}{|c|}{AE}        & 0.87 & 0.91 & 0.93 & 0.94 \\ \hline
    \multicolumn{1}{|c|}{\segmap{}} & 0.86 & 0.89 & 0.91 & 0.92 \\ \hline
  \end{tabular}
  \vspace{-4mm}
\end{table}

\subsubsection{Multi-robot SLAM in urban scenario}
\label{ssec_kitti_slam}
In order to simulate a multi-robot setup, we split sequence 00 of the KITTI odometry dataset into five sequences which are simultaneously played back on a single computer for a duration of 114 seconds.
%
In this experiment, the semantic information extracted from the \segmap{} descriptors is used to reject segments classified as \textit{vehicles} from the retrieval process.

With this setup, $113$ global associations were discovered, allowing to link all the robot trajectories and create a common representation.
We note that performing \ac{ICP} between the associated point clouds would refine the localization transformation by, on average, only \unit{0.13\pm0.06}{\meter} which is in the order of our voxelization resolution. 
However, this would require the original point cloud data to be kept in memory and transmitted to the central computer. 
Future work could consider refining the transformations by performing \ac{ICP} on the reconstructions.

Localization and map reconstruction was performed at an average frequency of \unit{10.5}{\hertz} and segment description was responsible for $30\%$ of this computing share with an average duration of \unit{28.4}{\milli\second} per local cloud (\unit{1.6}{\milli\second} per segment).
%
%
A section of the target map which has been reconstructed from the descriptors is depicted in Fig.~\ref{fig_teaser}.

Table~\ref{tbl:stats_experiment} presents the results of this experiment.
The required bandwidth is estimated by considering that each point is defined by three 32-bit floats and that 288 additional bits are required to link each descriptor to the trajectories.
We only consider the \textit{useful data} and ignore any transfer overheads.
The final map of the KITTI sequence 00 contains $1341$ segments out of which $284$ were classified as vehicles.
A map composed of all the raw segment point clouds would be $16.8$ MB whereas using our descriptor it is reduced to only $386.2$ kB.
This compression ratio of $43.5$x can be increased to $55.2$x if one decides to remove vehicles from the map.
%
%
This shows that our approach can be used for mapping much larger environments. 

\begin{table}
\fontsize{6}{8}\selectfont
  \renewcommand{\arraystretch}{1.2}
  \centering
  \caption{Statistics resulting from the three experiments.}
  \label{tbl:stats_experiment}
  \resizebox{1.0\columnwidth}{!}{%
  \begin{tabular}{|l|ccc|} \hline
    \textbf{Statistic} & \textbf{KITTI} & \textbf{Powerplant} & \textbf{Foundry} \\ \hline
    Duration (s)  & 114 & 850  & 1086 \\ \hline
    Number of robots  & 5 & 3  & 2 \\ \hline
    Number of segmented local cloud  & 557 & 758  & 672 \\ \hline
    Average  number of segments per cloud & 42.9 & 37.0  & 45.4 \\ \hline
    Bandwidth for transmitting local clouds (kB/s) & 4814.7 & 1269.2  & 738.1 \\ \hline
	Bandwidth for transmitting segments (kB/s) & 2626.6 & 219.4  & 172.2 \\ \hline    
    Bandwidth for transmitting descriptors (kB/s) & 60.4 & 9.5  & 8.1 \\ \hline
    Final map size with the \segmap{} descriptor (kB) & 386.2 & 181.3  & 121.2 \\ \hline
    Number of successful localizations & 113 & 27 & 85 \\ \hline
  \end{tabular}}
  \vspace{-5mm}
\end{table}


\subsubsection{Multi-robot SLAM in disaster environments}
\begin{figure}
  \centering
  \includegraphics[width=1.0\columnwidth]{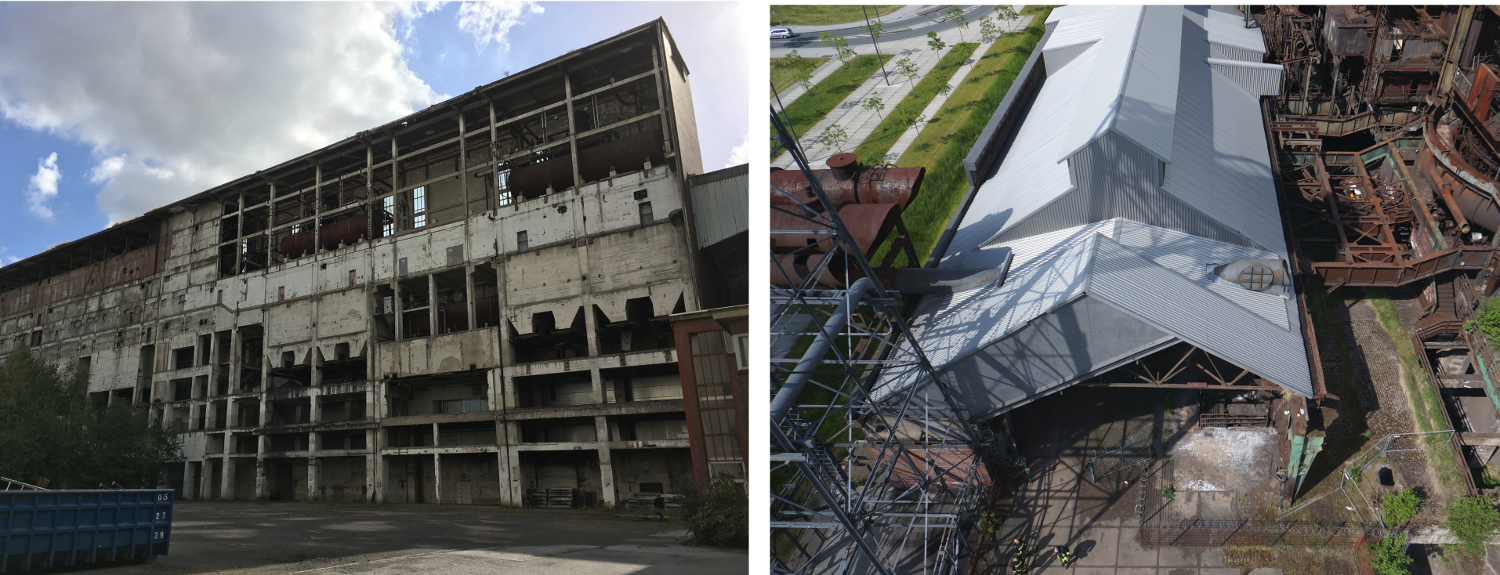}
  \vspace{-5mm}
  \caption{Buildings of the Gustav Knepper powerplant (left) and the Phoenix-West foundry (right).}
  \label{fig:environments}
  \vspace{-1mm}
\end{figure}

For the two following experiments, we use data collected by \acp{UGV} equipped with multiple motor encoders, an Xsens MTI-G \ac{IMU} and a rotating 2D SICK LMS-151 LiDAR.
First, three \acp{UGV} were deployed at the decommissioned Gustav Knepper powerplant: a large two-floors utility building measuring \unit{100}{\meter} long by \unit{25}{\meter} wide.
The second mission took place at the Phoenix-West foundry in a semi-open building made of steel.
A section measuring \unit{100}{\meter} by \unit{40}{\meter} was mapped using two \acp{UGV}.
The buildings are shown in Fig~\ref{fig:environments}.
%

\begin{figure}
  \centering
  \includegraphics[width=1.0\columnwidth]{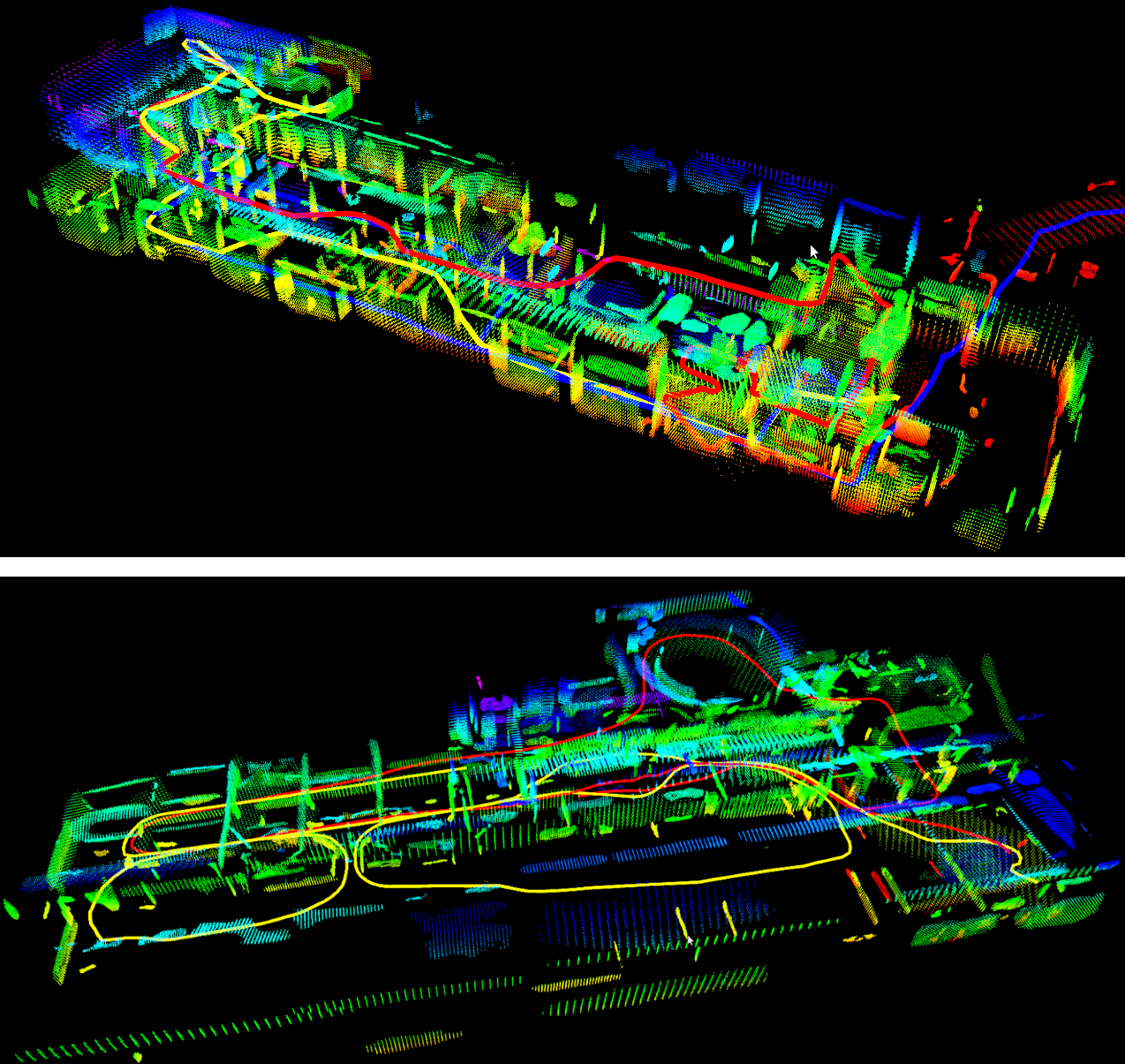}
  \vspace{-5mm}
  \caption{This figure illustrates a reconstruction of the buildings of the Gustav Knepper powerplant (top) and the Phoenix-West foundry (bottom).
  The point clouds are colored by height and the estimated robot trajectories are depicted with colored lines.
  }
  \label{fig:reconstruction_sar}
  \vspace{-6mm}
\end{figure}

For these two experiments, we used an incremental smoothness based region growing algorithm which extracts plane-like segments~\cite{dube2018incremental}.
The resulting \segmap{} reconstructions are shown in  Fig.~\ref{fig:reconstruction_sar} and detailed statistics are  presented in Table~\ref{tbl:stats_experiment}.
Although these planar segments have a very different nature than the ones used for training the descriptor extractor, multiple localizations have been made in real-time so that consistent maps  could be reconstructed in both experiments. 

\section{CONCLUSION}
\label{sec:conclusion}
This paper presented \segmap{}: a segment based approach for \textit{map representation} in localization and mapping with 3D sensors.
In essence, the robot's surroundings are decomposed into a set of segments, and each segment is represented by a distinctive, low dimensional learning based descriptor.
Data associations are identified by segment descriptor retrieval and matching, made possible by the repeatable and descriptive nature of segment based features.
The descriptive power of \segmap{} outperforms hand-crafted features as well as the evaluated autoencoder baseline.

In addition to enabling global localization, the \segmap{} descriptor allows us to reconstruct a map of the environment and to extract semantic information.
The ability to reconstruct the environment while achieving a high compression rate is one of the main features of \segmap{}.
This feature allows performing \ac{SLAM} with 3D LiDARs at a large scale requiring low communication bandwidth between the robots and a central computer.
These capabilities have been demonstrated through experiments with real-world data in urban driving and search and rescue scenarios.
The reconstructed maps could allow performing navigation tasks such as, for instance, multi-robot global path planning or increasing situational awareness.

In future work, we would like to extend the \segmap{} approach to different sensor modalities and different point cloud segmentation algorithms.
Furthermore, whereas the present work performs segment description in a discrete manner, it would be interesting to investigate incremental updates of learning based descriptors that could make the description process more efficient, such as the voting scheme proposed by~\citet{engelcke2017vote3deep}.
Moreover, it could of interest to learn the usefulness of segments as a precursory step to localization, based on their distinctiveness and semantic attributes.

\section*{ACKNOWLEDGMENTS}
This work was supported by the European Union's Seventh Framework Programme for research, technological development and demonstration under the TRADR project No. FP7-ICT-609763 and by the National Center of Competence in Research (NCCR) Robotics through the Swiss National Science Foundation.
The authors would like to thank Hannes Sommer, Mark Pfeiffer, Mattia Gollub, Helen Oleynikova, Abel Gawel,  Dr. Philipp Kr\"usi, Dr. Elena Stumm, and Alexander Winkler for their valuable collaboration and support. 


\bibliographystyle{plainnat}
\bibliography{rss18.bib}

\begin{thebibliography}{39}
\providecommand{\natexlab}[1]{#1}
\providecommand{\url}[1]{\texttt{#1}}
\expandafter\ifx\csname urlstyle\endcsname\relax
  \providecommand{\doi}[1]{doi: #1}\else
  \providecommand{\doi}{doi: \begingroup \urlstyle{rm}\Url}\fi

\bibitem[Bosse and Zlot(2009)]{Bosse2009a}
Michael Bosse and Robert Zlot.
\newblock {Continuous 3D Scan-Matching with a Spinning 2D Laser}.
\newblock In \emph{IEEE International Conference on Robotics and Automation},
  pages 4312--4319, 2009.

\bibitem[Brock et~al.(2016)Brock, Lim, Ritchie, and
  Weston]{brock2016generative}
Andrew Brock, Theodore Lim, JM~Ritchie, and Nick Weston.
\newblock {Generative and Discriminative Voxel Modeling with Convolutional
  Neural Networks}.
\newblock In \emph{Workshop on {3D} Deep Learning, NIPS}, 2016.

\bibitem[Bromley et~al.(1994)Bromley, Guyon, LeCun, S{\"a}ckinger, and
  Shah]{bromley1994signature}
Jane Bromley, Isabelle Guyon, Yann LeCun, Eduard S{\"a}ckinger, and Roopak
  Shah.
\newblock {Signature Verification using a "Siamese" Time Delay Neural Network}.
\newblock In \emph{Advances in Neural Information Processing Systems}, pages
  737--744, 1994.

\bibitem[Cadena et~al.(2016)Cadena, Carlone, Carrillo, Latif, Scaramuzza,
  Neira, Reid, and Leonard]{Cadena16tro-SLAMfuture}
C.~Cadena, L.~Carlone, H.~Carrillo, Y.~Latif, D.~Scaramuzza, J.~Neira, I.~Reid,
  and J.J. Leonard.
\newblock {Past, Present, and Future of Simultaneous Localization And Mapping:
  Towards the Robust-Perception Age}.
\newblock \emph{{IEEE} Transactions on Robotics}, 32\penalty0 (6):\penalty0
  1309--1332, 2016.

\bibitem[Cramariuc et~al.(2017)Cramariuc, Dub\'e, Sommer, Siegwart, and
  Gilitschenski]{cramariuc2017}
Andrei Cramariuc, Renaud Dub\'e, Hannes Sommer, Roland Siegwart, and Igor
  Gilitschenski.
\newblock {Learning 3D Segment Descriptors for Place Recognition}.
\newblock In \emph{Workshop on Learning for Localization and Mapping, IROS},
  2017.

\bibitem[Dai et~al.(2017)Dai, Ruizhongtai~Qi, and Niessner]{dai2016shape}
Angela Dai, Charles Ruizhongtai~Qi, and Matthias Niessner.
\newblock {Shape Completion Using 3D-Encoder-Predictor CNNs and Shape
  Synthesis}.
\newblock In \emph{{IEEE} Conference on Computer Vision and Pattern
  Recognition}, July 2017.

\bibitem[Dub{\'e} et~al.(2017{\natexlab{a}})Dub{\'e}, Dugas, Stumm, Nieto,
  Siegwart, and Cadena]{dube2017segmatch}
Renaud Dub{\'e}, Daniel Dugas, Elena Stumm, Juan Nieto, Roland Siegwart, and
  Cesar Cadena.
\newblock {SegMatch}: Segment based place recognition in {3D} point clouds.
\newblock In \emph{{IEEE} International Conference on Robotics and Automation},
  pages 5266--5272, 2017{\natexlab{a}}.

\bibitem[Dub{\'e} et~al.(2017{\natexlab{b}})Dub{\'e}, Gawel, Sommer, Nieto,
  Siegwart, and Cadena]{dube2017multirobot}
Renaud Dub{\'e}, Abel Gawel, Hannes Sommer, Juan Nieto, Roland Siegwart, and
  Cesar Cadena.
\newblock {An Online Multi-Robot SLAM System for 3D LIDARs}.
\newblock In \emph{{IEEE} International Conference on Robotics and Automation},
  pages 1004--1011, 2017{\natexlab{b}}.

\bibitem[Dub\'e et~al.(2018)Dub\'e, Gollub, Sommer, Igilitschenski, Siegwart,
  Cadena, and Nieto]{dube2018incremental}
Renaud Dub\'e, Mattia Gollub, Hannes Sommer, Igor Igilitschenski, Roland
  Siegwart, Cesar Cadena, and Juan Nieto.
\newblock {Incremental Segment-Based Localization in 3D Point Clouds}.
\newblock \emph{IEEE Robotics and Automation Letters}, 3\penalty0 (1):\penalty0
  1--8, 2018.

\bibitem[Elbaz et~al.(2017)Elbaz, Avraham, and Fischer]{elbaz20173d}
Gil Elbaz, Tamar Avraham, and Anath Fischer.
\newblock {3D Point Cloud Registration for Localization using a Deep Neural
  Network Auto-Encoder}.
\newblock In \emph{{IEEE} Conference on Computer Vision and Pattern
  Recognition}, pages 2472--2481. IEEE, 2017.

\bibitem[Elseberg et~al.(2012)Elseberg, Magnenat, Siegwart, and
  N{\"u}chter]{elsebergcomparison}
J.~Elseberg, S.~Magnenat, R.~Siegwart, and A.~N{\"u}chter.
\newblock {Comparison of Nearest-Neighbor-Search Strategies and Implementations
  for Efficient Shape Registration}.
\newblock \emph{Journal of Software Engineering for Robotics}, 3\penalty0
  (1):\penalty0 2--12, 2012.
\newblock ISSN 2035-3928.

\bibitem[Engelcke et~al.(2017)Engelcke, Rao, Wang, Tong, and
  Posner]{engelcke2017vote3deep}
Martin Engelcke, Dushyant Rao, Dominic~Zeng Wang, Chi~Hay Tong, and Ingmar
  Posner.
\newblock {Vote3Deep: Fast Object Detection in 3D Point Clouds Using Efficient
  Convolutional Neural Networks}.
\newblock In \emph{{IEEE} International Conference on Robotics and Automation},
  pages 1355--1361. IEEE, 2017.

\bibitem[Geiger et~al.(2012)Geiger, Lenz, and Urtasun]{geiger2012we}
Andreas Geiger, Philip Lenz, and Raquel Urtasun.
\newblock {Are we ready for Autonomous Driving? The KITTI Vision Benchmark
  Suite}.
\newblock In \emph{{IEEE} Conference on Computer Vision and Pattern
  Recognition}, 2012.

\bibitem[Glorot and Bengio(2010)]{glorot2010understanding}
Xavier Glorot and Yoshua Bengio.
\newblock Understanding the difficulty of training deep feedforward neural
  networks.
\newblock In \emph{International Conference on Artificial Intelligence and
  Statistics}, volume~9, pages 249--256, 2010.

\bibitem[Guizilini and Ramos(2017)]{guizilini2017learning}
Vitor Guizilini and Fabio Ramos.
\newblock {Learning to Reconstruct 3D Structures for Occupancy Mapping}.
\newblock In \emph{Robotics: Science and Systems}, 2017.

\bibitem[Ioffe and Szegedy(2015)]{ioffe2015batch}
Sergey Ioffe and Christian Szegedy.
\newblock {Batch normalization: Accelerating deep network training by reducing
  internal covariate shift}.
\newblock In \emph{International Conference on Machine Learning}, pages
  448--456, 2015.

\bibitem[Kaess et~al.(2012)Kaess, Johannsson, Roberts, Ila, Leonard, and
  Dellaert]{Kaess2012a}
M.~Kaess, H.~Johannsson, R.~Roberts, V.~Ila, J.~J. Leonard, and F.~Dellaert.
\newblock {iSAM2: Incremental smoothing and mapping using the Bayes tree}.
\newblock \emph{The International Journal of Robotics Research}, 31\penalty0
  (2):\penalty0 216--235, 2012.

\bibitem[Krizhevsky et~al.(2012)Krizhevsky, Sutskever, and
  Hinton]{krizhevsky2012imagenet}
Alex Krizhevsky, Ilya Sutskever, and Geoffrey~E Hinton.
\newblock Imagenet classification with deep convolutional neural networks.
\newblock In \emph{Advances in Neural Information Processing Systems}, pages
  1097--1105, 2012.

\bibitem[Li et~al.(2016)Li, Zhang, and Xia]{livehicle}
Bo~Li, Tianlei Zhang, and Tian Xia.
\newblock {Vehicle Detection from 3D Lidar Using Fully Convolutional Network}.
\newblock In \emph{Robotics: Science and Systems}, 2016.

\bibitem[Lowry et~al.(2016)Lowry, Sunderhauf, Newman, Leonard, Cox, Corke, and
  Milford]{lowryvisual}
Stephanie Lowry, Niko Sunderhauf, Paul Newman, John~J Leonard, David Cox, Peter
  Corke, and Michael~J Milford.
\newblock {Visual Place Recognition: A Survey}.
\newblock \emph{{IEEE} Transactions on Robotics}, 2016.

\bibitem[Maturana and Scherer(2015)]{maturana2015voxnet}
Daniel Maturana and Sebastian Scherer.
\newblock {VoxNet: A 3D Convolutional Neural Network for real-time object
  recognition}.
\newblock In \emph{{IEEE} International Conference on Robotics and Automation},
  2015.

\bibitem[Newman et~al.(2009)Newman, Sibley, Smith, Cummins, Harrison, Mei,
  Posner, Shade, Schroeter, Murphy, et~al.]{newman2009navigating}
Paul Newman, Gabe Sibley, Mike Smith, Mark Cummins, Alastair Harrison, Chris
  Mei, Ingmar Posner, Robbie Shade, Derik Schroeter, Liz Murphy, et~al.
\newblock {Navigating, Recognizing and Describing Urban Spaces With Vision and
  Lasers}.
\newblock \emph{The International Journal of Robotics Research}, 28\penalty0
  (11-12):\penalty0 1406--1433, 2009.

\bibitem[P. and L.(2015)]{adam}
Kingma~D. P. and Ba~J. L.
\newblock {Adam: A Method for Stochastic Optimization}.
\newblock In \emph{International Conference on Learning Representations,
  1–13}, 2015.

\bibitem[Parkhi et~al.(2015)Parkhi, Vedaldi, Zisserman, et~al.]{parkhi2015deep}
Omkar~M Parkhi, Andrea Vedaldi, Andrew Zisserman, et~al.
\newblock {Deep Face Recognition}.
\newblock In \emph{British Machine Vision Conference}, volume~1, page~6, 2015.

\bibitem[Qi et~al.(2017)Qi, Su, Mo, and Guibas]{qi2016pointnet}
Charles~R. Qi, Hao Su, Kaichun Mo, and Leonidas~J. Guibas.
\newblock {PointNet: Deep Learning on Point Sets for 3D Classification and
  Segmentation}.
\newblock In \emph{{IEEE} Conference on Computer Vision and Pattern
  Recognition}, July 2017.

\bibitem[Ricao~Canelhas et~al.(2017)Ricao~Canelhas, Schaffernicht, Stoyanov,
  Lilienthal, and Davison]{ricao2017compressed}
Daniel Ricao~Canelhas, Erik Schaffernicht, Todor Stoyanov, Achim~J Lilienthal,
  and Andrew~J Davison.
\newblock {Compressed Voxel-Based Mapping Using Unsupervised Learning}.
\newblock \emph{Robotics}, 6\penalty0 (3):\penalty0 15, 2017.

\bibitem[Riegler et~al.(2017)Riegler, Osman~Ulusoy, and
  Geiger]{riegler2016octnet}
Gernot Riegler, Ali Osman~Ulusoy, and Andreas Geiger.
\newblock {OctNet: Learning Deep 3D Representations at High Resolutions}.
\newblock In \emph{{IEEE} Conference on Computer Vision and Pattern
  Recognition}, 2017.

\bibitem[Sch{\"o}nberger et~al.(2018)Sch{\"o}nberger, Pollefeys, Geiger, and
  Sattler]{schonberger2017semantic}
Johannes~L Sch{\"o}nberger, Marc Pollefeys, Andreas Geiger, and Torsten
  Sattler.
\newblock {Semantic Visual Localization}.
\newblock 2018.

\bibitem[Srivastava et~al.(2014)Srivastava, Hinton, Krizhevsky, Sutskever, and
  Salakhutdinov]{srivastava2014dropout}
Nitish Srivastava, Geoffrey~E Hinton, Alex Krizhevsky, Ilya Sutskever, and
  Ruslan Salakhutdinov.
\newblock {Dropout: A Simple Way to Prevent Neural Networks from Overfitting}.
\newblock \emph{Journal of machine learning research}, 15\penalty0
  (1):\penalty0 1929--1958, 2014.

\bibitem[Tchapmi et~al.(2017)Tchapmi, Choy, Armeni, Gwak, and
  Savarese]{tchapmi2017segcloud}
Lyne~P. Tchapmi, Christopher~B. Choy, Iro Armeni, JunYoung Gwak, and Silvio
  Savarese.
\newblock {SEGCloud: Semantic Segmentation of 3D Point Clouds}.
\newblock In \emph{International Conference on 3D Vision}, 2017.

\bibitem[Varley et~al.(2017)Varley, DeChant, Richardson, Ruales, and
  Allen]{varley2016shape}
J.~Varley, C.~DeChant, A.~Richardson, J.~Ruales, and P.~Allen.
\newblock {Shape Completion Enabled Robotic Grasping}.
\newblock In \emph{{IEEE} International Conference on Robotics and Automation},
  pages 2442--2447, 2017.

\bibitem[Weinberger et~al.(2006)Weinberger, Blitzer, and
  Saul]{weinberger2006distance}
Kilian~Q Weinberger, John Blitzer, and Lawrence~K Saul.
\newblock {Distance Metric Learning for Large Margin Nearest Neighbor
  Classification}.
\newblock In \emph{Advances in neural information processing systems}, pages
  1473--1480, 2006.

\bibitem[Weinmann et~al.(2014)Weinmann, Jutzi, and
  Mallet]{weinmann2014semantic}
Martin Weinmann, Boris Jutzi, and Cl{\'e}ment Mallet.
\newblock {Semantic 3D scene interpretation: A framework combining optimal
  neighborhood size selection with relevant features}.
\newblock \emph{ISPRS Annals of the Photogrammetry, Remote Sensing and Spatial
  Information Sciences}, 2\penalty0 (3):\penalty0 181, 2014.

\bibitem[Wohlhart and Lepetit(2015)]{wohlhart2015learning}
Paul Wohlhart and Vincent Lepetit.
\newblock {Learning Descriptors for Object Recognition and 3D Pose Estimation}.
\newblock In \emph{{IEEE} Conference on Computer Vision and Pattern
  Recognition}, pages 3109--3118, 2015.

\bibitem[Wohlkinger and Vincze(2011)]{wohlkinger2011ensemble}
Walter Wohlkinger and Markus Vincze.
\newblock {Ensemble of shape functions for 3D object classification}.
\newblock In \emph{{IEEE} International Conference on Robotics and
  Biomimetics}, 2011.

\bibitem[Wu et~al.(2016)Wu, Zhang, Xue, Freeman, and Tenenbaum]{wu2016learning}
Jiajun Wu, Chengkai Zhang, Tianfan Xue, Bill Freeman, and Josh Tenenbaum.
\newblock {Learning a Probabilistic Latent Space of Object Shapes via 3D
  Generative-Adversarial Modeling}.
\newblock In \emph{Advances in Neural Information Processing Systems}, pages
  82--90, 2016.

\bibitem[Wu et~al.(2015)Wu, Song, Khosla, Yu, Zhang, Tang, and Xiao]{wu20153d}
Zhirong Wu, Shuran Song, Aditya Khosla, Fisher Yu, Linguang Zhang, Xiaoou Tang,
  and Jianxiong Xiao.
\newblock {3D ShapeNets: A Deep Representation for Volumetric Shapes}.
\newblock In \emph{{IEEE} Conference on Computer Vision and Pattern
  Recognition}, pages 1912--1920, 2015.

\bibitem[Zeng et~al.(2017)Zeng, Song, Nie{\ss}ner, Fisher, Xiao, and
  Funkhouser]{zeng20163dmatch}
Andy Zeng, Shuran Song, Matthias Nie{\ss}ner, Matthew Fisher, Jianxiong Xiao,
  and Thomas Funkhouser.
\newblock {3DMatch: Learning Local Geometric Descriptors from RGB-D
  Reconstructions}.
\newblock In \emph{{IEEE} Conference on Computer Vision and Pattern
  Recognition}, 2017.

\bibitem[Zhang and Singh(2014)]{zhang2014loam}
Ji~Zhang and Sanjiv Singh.
\newblock {LOAM: Lidar Odometry and Mapping in Real-time}.
\newblock In \emph{Robotics: Science and Systems}, 2014.

\end{thebibliography}

\acrodefplural{CNN}[CNNs]{Convolutional Neural Networks}
\acrodefplural{UGV}[UGVs]{Unmanned Ground Vehicles}
\acrodefplural{ReLU}[ReLUs]{Rectified Linear Units}

\begin{acronym}
\acro{ICP}{Iterative Closest Point}
\acro{MAP}{Maximum A Posteriori}
\acro{SLAM}{Simultaneous Localization and Mapping}
\acro{DoF}{Degrees of Freedom}
\acro{GUI}{Graphical User Interface}
\acro{TRADR}{``Long-Term Human-Robot Teaming for Robots Assisted Disaster Response''}
\acro{SaR}{Search and Rescue}
\acro{UGV}{Unmanned Ground Vehicle}
\acro{IMU}{Inertial Measurement Unit}
\acro{k-NN}{k-Nearest Neighbors}
\acro{FPFH}{Fast Point Feature Histograms}
\acro{CNN}{Convolutional Neural Network}
\acro{ReLU}{Rectified Linear Unit}
\acro{ROC}{Receiver Operating Characteristic}
\acro{ADAM}{Adaptive Moment Estimation}
\acro{SGD}{Stochastic Gradient Descent}
\acro{TPR}{True Positive Rate}
\acro{FPR}{False Positive Rate}
\end{acronym}

\end{document}